\documentclass[sigconf]{acmart}
\AtBeginDocument{%
  }

\copyrightyear{2026}
\acmYear{2026}
\setcopyright{cc}
\setcctype{by}

\acmConference[MM '26]
{Proceedings of the 34th ACM International Conference on Multimedia}
{November 10--14, 2026}
{Rio de Janeiro, Brazil.}

\acmBooktitle{Proceedings of the 34th ACM International Conference on Multimedia
(MM '26), November 10--14, 2026, Rio de Janeiro, Brazil}

\acmISBN{979-8-4007-2213-4/2026/11}
\acmDOI{10.1145/3767308.3837689}

\usepackage{xspace}
\usepackage{enumitem}
\usepackage{multirow}
\usepackage{makecell}
\usepackage{colortbl}
\usepackage{placeins}

\makeatletter

\dimen\footins=8in
\pretocmd{\@mkabstract}{\clearpage}{}{}

\DeclareRobustCommand\onedot{\futurelet\@let@token\@onedot}
\def\@onedot{\ifx\@let@token.\else.\null\fi\xspace}

\def\ie{\emph{i.e}\onedot}

\makeatother

\begin{document}

\title{Chronos: Sequential Identity-Preserving Video Generation by Temporal Anchoring}

\author{Mengqi Huang}
\authornote{Equal contribution.}
\affiliation{%
  \institution{University of Science and Technology of China}
  \city{Hefei}
  \state{Anhui}
  \country{China}}
\email{huangmq@ustc.edu.cn}

\author{Wenchuan Wang}
\authornotemark[1]
\affiliation{%
  \institution{University of Science and Technology of China}
  \city{Hefei}
  \state{Anhui}
  \country{China}}
\email{wenc\_k@mail.ustc.edu.cn}

\author{Zhendong Mao}
\authornote{Corresponding author.}
\affiliation{%
  \institution{University of Science and Technology of China}
  \city{Hefei}
  \state{Anhui}
  \country{China}}
\email{zdmao@ustc.edu.cn}

\author{Yijing Tu}
\affiliation{%
  \institution{University of Science and Technology of China}
  \city{Hefei}
  \state{Anhui}
  \country{China}}
\email{tuyijing@mail.ustc.edu.cn}

\author{Zhike Liu}
\affiliation{%
  \institution{University of Science and Technology of China}
  \city{Hefei}
  \state{Anhui}
  \country{China}}
\email{liuzhike@mail.ustc.edu.cn}

\author{Luoni Wang}
\affiliation{%
  \institution{University of Science and Technology of China}
  \city{Hefei}
  \state{Anhui}
  \country{China}}
\email{wangluoni@mail.ustc.edu.cn}

\author{Ning Tong}
\affiliation{%
  \institution{South China University of Technology}
  \city{Guangzhou}
  \state{Guangdong}
  \country{China}}
\email{202364810141@mail.scut.edu.cn}

\author{Xiaoxiao Sheng}
\affiliation{%
  \institution{Huawei Inc.}
  \city{Shanghai}
  \country{China}}
\email{shengxiaoxiao@sjtu.edu.cn}

\author{Tianyu Huang}
\affiliation{%
  \institution{Huawei Inc.}
  \city{Shanghai}
  \country{China}}
\email{tyhuang0428@gmail.com}

\author{Xuecheng Qi}
\affiliation{%
  \institution{Independent Researcher}
  \city{Hangzhou}
  \state{Zhejiang}
  \country{China}}
\email{xuechengqi7@gmail.com}

\author{Yongdong Zhang}
\affiliation{%
  \institution{University of Science and Technology of China}
  \city{Hefei}
  \state{Anhui}
  \country{China}}
\email{zhyd73@ustc.edu.cn}

\renewcommand{\shortauthors}{Mengqi Huang et al.}

\begin{abstract}
Identity-preserving video generation aims to synthesize high-quality and visually realistic videos from textual and visual conditions while faithfully maintaining the subject's identity across diverse motions. Prior methods have extensively introduced reference-guided control signals and identity-aware conditioning to improve facial alignment from reference images. However, improving static identity alignment alone creates a fundamental tension between identity fidelity and temporal controllability, which frequently causes identity drift or motion suppression under large pose variations and multi-segment sequential actions. To address this, we propose \textbf{\textit{Chronos}}, a framework that unifies single-segment facial identity preservation and multi-segment sequential action generation under a common temporal formulation. Specifically, \textit{Chronos} integrates \textbf{Semantic Prompt Anchoring} to adaptively refine the user prompt, \textbf{Visual Frame Anchoring} to transform conditions into visually grounded keyframe anchors for visual guidance, and \textbf{Metric-Guided Candidate Routing} to select the highest-quality videos from heterogeneous generators via multi-dimensional evaluation. Extensive evaluations demonstrate the superiority of our approach, which \textbf{achieved first place in both Track 1 and Track 2 of the ACM MM 2026 Identity-Preserving Video Generation Challenge}, demonstrating its effectiveness across both facial identity preservation and sequential action generation settings.
\end{abstract}

\ccsdesc[500]{Computing methodologies~Computational photography}

\keywords{Video Generation; Diffusion Transformer; AIGC; Multi-Modal Generation; Identity-Preserving Video Generation}

\begin{teaserfigure}
  \setlength{\abovecaptionskip}{0pt}
  \centering
  \includegraphics[width=.79\textwidth]{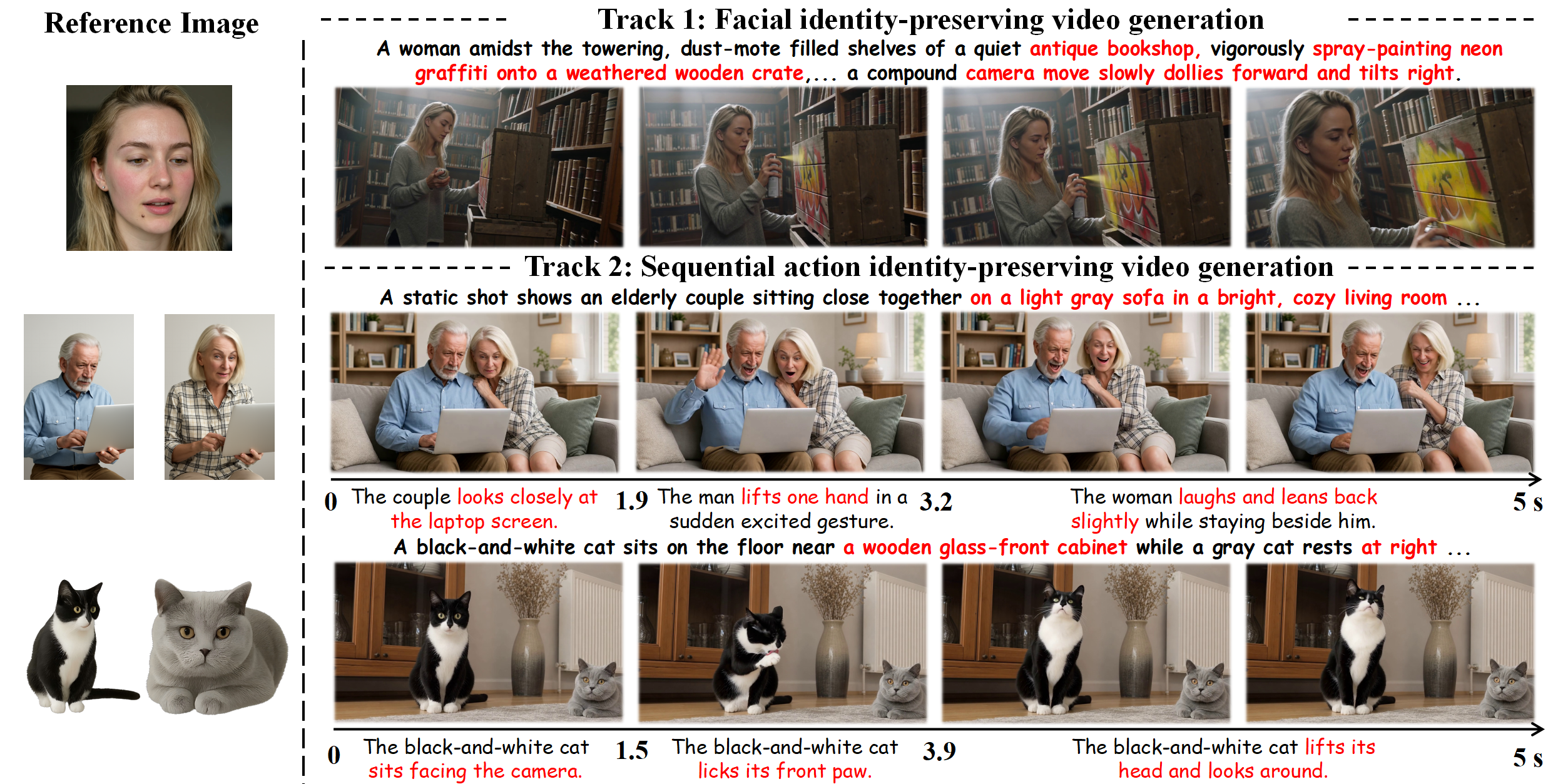}
  \caption{Generated results of our proposed \textit{Chronos}. \textit{Chronos} enables multi-subject video generation with precise and flexible temporal control. Given reference images, a raw prompt, and optional temporal captions, \textit{Chronos} generates text-aligned, high-quality videos with high identity fidelity. \textcolor{red}{Red} text highlights key elements in long instructions.}
  \Description{Reference images and generated video frames for facial identity preservation and sequential multi-subject action generation. Red phrases indicate temporally controlled actions in the prompts.}
  \label{fig:1}
\end{teaserfigure}

\maketitle

\section{Introduction}
Recent advances in generative modeling~\cite{scaling2024, wan2025wan, hunyuanvideo2024,lin2025realgeneral,wang2025dualreal,tu2026stream,wu2026stream,fu2026lance,mao2025realcustom++,zheng2026pro3d,jia2025moga,chen2025longanimation,tu2025a4a} have substantially improved visual realism, controllability, and generation fidelity across diverse visual domains.
In video generation, modern Text-to-Video (T2V)~\cite{ho2022video,hong2022cogvideo,yang2025cogvideox,singer2022make,ho2022imagen}, Image-to-Video (I2V)~\cite{chen2023videocrafter1,xing2024dynamicrafter,zhang2023i2vgen,blattmann2023stable}, and First-Last-Frame-to-Video (FLF2V)~\cite{jiang2018super,bao2019depth,chen2023seine,guo2024sparsectrl,jain2024video} models can synthesize high-quality videos from diverse textual and visual conditions, enabling applications in content creation, virtual character generation, and personalized media.
Nevertheless, faithful subject identity preservation remains challenging, particularly under large pose variations, non-frontal viewpoints, occlusions, and multi-subject interactions, where the reference identity can be easily weakened or confused across frames.

Prior methods have introduced identity-aware conditioning from reference inputs, ranging from frequency-domain facial representations in ConsistID~\cite{consist-id2025} to joint identity-semantic modeling in PersonalVideo~\cite{personal-video2025}. Recent reference-guided methods, such as RefAlign~\cite{refalign2026}, Phantom~\cite{phantom2025}, Stand-In~\cite{stand-in2025}, and VACE~\cite{vace2025}, further validate reference images as an effective control signal for identity preservation.
However, improving reference-based identity alignment alone is insufficient for video generation, where identity must be preserved jointly with motion, scene composition, and temporal progression.
This creates a fundamental tension between identity fidelity and temporal controllability, \ie, overly strong identity constraints can suppress motion diversity, restrict camera freedom, and introduce temporal rigidity, while weak identity conditioning often causes identity drift under small faces, occlusions, non-frontal views, or action-induced appearance changes.
The challenge becomes more pronounced in multi-subject and sequential-action settings, where the model must preserve distinct identities while maintaining coherent transitions across multiple temporal phases.

To address these challenges, we propose \textbf{\textit{Chronos}}, a training-free framework for identity-preserving video generation. It unifies facial identity preservation and sequential-action generation under a common temporal formulation.
Technically, given reference images, a raw prompt, and optional temporal captions, \textit{Chronos} first performs \textbf{Semantic Prompt Anchoring} to extract compact, reliable identity cues from the reference images and conservatively refines the prompt.
Instead of over-specifying appearance details, this module adaptively adjusts visibility and viewpoint cues based on the estimated generation risk, improving identity preservation while maintaining the original scene semantics and motion intent.
We further introduce \textbf{Visual Frame Anchoring}, which transforms the enhanced textual condition and reference identity into visually grounded frame anchors. In the single-segment setting, VFA generates a high-fidelity first-frame anchor to provide a stable initialization for image-to-video generation. In the sequential-action setting, it constructs temporally grounded key frames in a cascaded manner, enabling different action phases to be guided by identity-consistent visual anchors. By reducing the ambiguity of text-only conditioning, these anchors provide stronger visual guidance for challenging pose, viewpoint, and action transitions.
Finally, we introduce \textbf{Metric-Guided Candidate Routing} to exploit the complementary strengths of heterogeneous video generators. Given the enhanced prompts and generated visual anchors, the generator pool produces diverse candidate videos. An automatic evaluation system scores each candidate across multiple complementary dimensions. Based on these scores, the final output is selected through metric-guided routing, while an exception-aware review mechanism filters candidates with severe failures in critical dimensions before final acceptance.

Our contributions are summarized as follows:
\begin{itemize}[leftmargin=*, topsep=2pt, itemsep=2pt, parsep=0pt, partopsep=0pt]
    \item \textit{Chronos} unifies single-segment facial identity preservation and multi-segment sequential action generation under a common temporal anchoring formulation, providing a consistent framework for identity-preserving video generation across different temporal granularities.
    \item \textit{Chronos} employs \textbf{Semantic Prompt Anchoring} and \textbf{Visual Frame Anchoring} to bridge semantic and visual grounding gaps in the generation conditions, improving identity constraints and temporal controllability in a training-free manner. Meanwhile, the \textbf{Metric-Guided Candidate Routing} leverages heterogeneous video generators and multi-dimensional evaluation to select high-quality videos while filtering critical failure cases.
    \item \textit{Chronos} achieves \textbf{first place in both Track 1 and Track 2 of the ACM MM 2026 Identity-Preserving Video Generation Challenge}, demonstrating its effectiveness across both facial identity preservation and sequential action generation settings.
\end{itemize}

\section{Related Work}
\begin{figure*}[t]
    \centering
    \includegraphics[width=\linewidth]{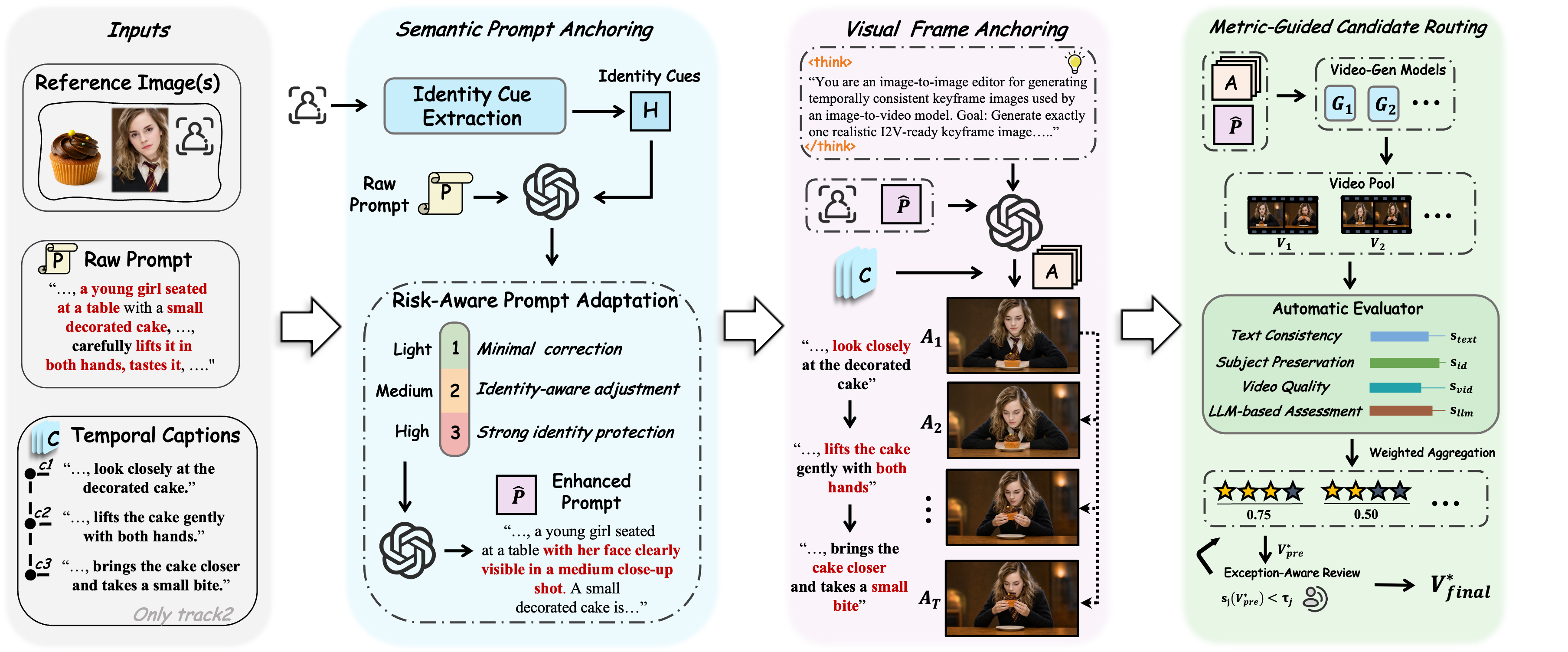}
    \caption{Our framework of \textit{Chronos}. Given reference images, a raw prompt, and optional temporal captions, the pipeline first performs Semantic Prompt Anchoring, then generates frame anchors through Visual Frame Anchoring, where Track 1 uses a single anchor $A_1$, and Track 2 may use additional segment-level anchors $\{A_t\}_{t=2}^{T}$. The enhanced prompt and anchors are passed to track-specific video generators, followed by metric-guided evaluation and automatic routing to obtain the final video.}
    \Description{A left-to-right four-stage flowchart. Reference images, a raw prompt, and optional temporal captions enter Semantic Prompt Anchoring, which extracts identity cues and applies one of three levels of risk-aware prompt adaptation. Visual Frame Anchoring then uses the enhanced prompt and temporal captions to create one or more ordered frame anchors. Multiple video generators produce candidate videos, which are scored for text consistency, subject preservation, video quality, and an LLM-based assessment before exception-aware review selects the final video.}
    \label{fig:pipeline}
\end{figure*}
\subsection{Identity-Preserving Video Generation}
In identity-preserving video generation~\cite{id-animator2024,consist-id2025,personal-video2025,phantom2025,vace2025,stand-in2025,wang2025dualreal}, balancing textual controllability with faithful identity preservation remains a fundamental challenge.
ID-Animator~\cite{id-animator2024} introduces a face adapter to encode identity-relevant facial attributes. ConsistID~\cite{consist-id2025} extracts frequency-domain facial features to construct identity representations, while PersonalVideo~\cite{personal-video2025} jointly models identity consistency and semantic consistency for personalized video generation. Despite notable progress in identity-preserving generation, maintaining rich and natural motion dynamics remains challenging, as identity constraints may often limit temporal variation in generated videos. To mitigate this issue, MagicID~\cite{magic-id2025} develops a dedicated framework for generating identity-consistent videos with enhanced motion dynamics.
More recently, Phantom~\cite{phantom2025} and VACE~\cite{vace2025} further highlight the promise of reference-guided video generation for controllable identity preservation. Moreover, Stand-In~\cite{stand-in2025} proposes a novel, lightweight adapter-free framework for identity-preserving generation, while RefAlign~\cite{refalign2026} further improves subject consistency and overall video quality. In the ACM MM 2025 Identity-Preserving Video Generation Challenge, TPIGE~\cite{tpige2025} and IPT2V~\cite{wang2025identity} present lightweight yet effective pipelines. 
We propose a training-free framework for identity-preserving
video generation, which unifies single-segment facial identity preservation and multi-segment sequential action generation under a common temporal anchoring formulation.

\subsection{Multi-Subject Identity Preservation}
Multi-subject identity preservation remains challenging, as generated videos must preserve the distinct identities of multiple subjects while maintaining temporal consistency across frames. Subject-to-Video (S2V) approaches such as VACE~\cite{vace2025} and Phantom~\cite{phantom2025} have advanced multi-subject identity preservation through reference-guided video generation. To alleviate the entanglement between foreground subjects and background regions, Kaleido~\cite{kaleido2025} introduces R-RoPE to handle background disentanglement. Nevertheless, existing approaches face challenges in maintaining temporal consistency while preserving subject identities. MotionCharacter~\cite{motion-character2026} introduces an effective framework for generating identity-preserving and motion-controllable videos. Furthermore, ID-Crafter~\cite{id-crafter2025} achieves competitive performance in multi-subject identity preservation and temporal consistency. 
To enhance multi-subject identity preservation and temporal consistency, we propose a keyframe-guided framework that fuses multi-reference information into identity-aware starting frames and leverages first-last-frame video generation to ensure seamless transitions between segments.

\section{Methodology}

\subsection{Overview}
Given reference images $\mathcal{I}$, a raw prompt $P$, and optional temporal captions $\mathcal{C}=\{c_t\}_{t=1}^{T}$, \textit{Chronos} aims to generate a video $V$ that maintains the target identity while remaining consistent with the specified scene, action, and temporal progression.
We unify \textbf{facial identity-preserving generation (Track 1)} and \textbf{sequential action identity-preserving generation (Track 2)} under a common temporal formulation, with Track 1 as the single-segment case ($T=1$) and Track 2 as the multi-segment case ($T>1$).

As detailed in ~\ref{fig:pipeline}, \textit{Chronos} consists of three stages: (1) we perform conservative \textbf{Semantic Prompt Anchoring} (SPA), which extracts compact, confident identity cues from the reference images and uses them to adaptively refine the raw prompt.
This design improves identity preservation without over-constraining the prompt, thereby maintaining camera diversity, motion amplitude, and the original scene semantics.
(2) we introduce \textbf{Visual Frame Anchoring} (VFA), which constructs visual anchors for downstream generation.
In Track 1, VFA produces a single first frame anchor $A_1$.
In Track 2, it further generates temporally grounded anchors $\{A_t\}_{t=1}^{T}$ to guide different action phases when additional visual constraints are needed.
(3) \textbf{Metric-Guided Candidate Routing} (MGCR) feeds the adapted textual conditions and generated visual anchors into a pool of video generators to obtain diverse candidates. These candidates are then evaluated across identity preservation, visual quality, and action-timeline alignment to select the best video.

\subsection{Semantic Prompt Anchoring}
To mitigate identity drift induced by challenging camera viewpoints, distant framing, or occlusions in the raw prompt $P$, we introduce a non-intrusive \textbf{Semantic Prompt Anchoring} (SPA) module.
Given the reference images $\mathcal{I}$ and the raw prompt $P$, the module refines the textual description to strengthen identity constraints while faithfully preserving the original scene semantics, style, and motion dynamics. This process consists of two steps:
\begin{itemize}
\item \textbf{Identity Cue Extraction}.
We first employ a GPT model~\cite{singh2025openai} conditioned on $\mathcal{I}$ to extract a compact identity representation $H$. To preserve the original scene semantics, this step discards transient attributes, such as clothing and background context, and distills only reliable, invariant biometric cues.
\item \textbf{Risk-Aware Prompt Adaptation.} 
Given $P$ and $H$, the model estimates potential risks in image-to-video generation and maps them to a discrete augmentation score $s \in \{1,2,3\}$, where larger values indicate stronger prompt intervention.
Then the model~\cite{singh2025openai} generates the enhanced prompt $\hat{P}$ by adding concise viewpoint and visibility cues under fidelity-preserving editing instruction.
\end{itemize}
By decoupling visual identity parsing from textual adaptation, this design provides a robust conditioning basis for downstream video synthesis, improving identity preservation while maintaining consistency with the original prompt.

\subsection{Visual Frame Anchoring}
Although the enhanced prompt $\hat{P}$ strengthens textual identity constraints, text-only conditioning remains insufficient when the target pose, action, or camera viewpoint substantially deviates from the reference image.
We therefore introduce \textbf{Visual Frame Anchoring} (VFA), a training-free visual anchoring mechanism that uses a high-quality image-to-image synthesis operator $\Phi_{\mathrm{I2I}}$ to produce visually grounded key frames from the reference identity and task-specific textual conditions.
Specifically, VFA takes three multimodal inputs: the reference images $\mathcal{I}$, the enhanced prompt $\hat{P}$, and a structured instruction template $S_{\mathrm{VFA}}$.
The anchoring workflow is adapted to the requirements of both evaluation tracks:
\begin{itemize}
    \item \textbf{Track 1:} The operator $\Phi_{\mathrm{I2I}}$ is executed with $\mathcal{I}$, $\hat{P}$, and $S_{\mathrm{VFA}}$ to synthesize a single first-frame anchor $A_1$. 
    \item \textbf{Track 2:} VFA further incorporates sequential motion captions $\mathcal{C}=\{c_t\}_{t=1}^{T}$. It first generates the initial anchor $A_1$ using the first caption $c_1$, the enhanced prompt, and the reference images. Subsequent anchors are then synthesized in a cascaded manner by reusing previously generated key frames as visual conditions.
\end{itemize}

Consequently, the generated anchors serve as visual guidance for downstream video generation, helping maintain identity preservation across complex pose, action, and behavioral transitions.
\subsection{Metric-Guided Candidate Routing}
\begin{table*}[t]
  \centering
  \caption{Challenge Results of the Top Three Teams. \textit{Chronos} achieves the first place in both Track
1 and Track 2. }
  \label{tab:challenge_results}
  \begin{tabular}{ccc@{\hspace{2em}}ccc}
    \toprule
    \multicolumn{3}{c}{\textbf{Track 1: Facial Identity-Preserving}} & 
    \multicolumn{3}{c}{\textbf{Track 2: Sequential Action Identity-Preserving}} \\
    \cmidrule(lr){1-3} \cmidrule(lr){4-6}
    \textbf{Team} & \textbf{Final Score$\downarrow$} & \textbf{Rank} & 
    \textbf{Team} & \textbf{Final Score$\downarrow$} & \textbf{Rank} \\
    \midrule
    \rowcolor{pink!50}\textbf{USTC-CMI} & 2.5 & 1 & \textbf{USTC-CMI} & 1.25   & 1 \\
    MIPL\_Video             & 2.5 & 1 & WislabGDUT              & 1.8125 & 2 \\
    xuanyuan\_fuxi          & 3   & 3 & USTC-AC                 & 2.9375 & 3 \\
    \bottomrule
  \end{tabular}
\end{table*}
\begin{table*}
    \centering
  \caption{Quantitative comparison of Different Methods on Track 1 and Track 2. \textit{Chronos} achieves the highest Total Score in both Track 1 and Track 2, demonstrating its superior overall performance. The best results are highlighted in bold.}
  \label{tab:main_results}
  \resizebox{\linewidth}{!}{
  \centering
  \begin{tabular}{ccccccccc}
    \toprule
    \multirow{2}{*}{\textbf{Method}} &
    \textbf{Identity Consistency} &
    \multicolumn{2}{c}{\textbf{Text Alignment}} &
    \multicolumn{4}{c}{\textbf{Video Quality}} &
    \multirow{2}{*}{\textbf{Total Score$\uparrow$}} \\
    \cmidrule(lr){2-2}  \cmidrule(lr){3-4} \cmidrule(lr){5-8} 
    & \textbf{\makecell{Subject Score$\uparrow$}}
    & \textbf{\makecell{GMEScore$\uparrow$}}
    & \textbf{\makecell{GMEScore\_Seg$\uparrow$}}
    & \textbf{\makecell{Aesthetic Score$\uparrow$}}
    & \textbf{\makecell{Motion Amplitude$\uparrow$}}
    & \textbf{\makecell{Motion Smoothness$\uparrow$}}
    & \textbf{\makecell{Natural Score$\uparrow$}} & \\
    \midrule
    
    \rowcolor{gray!15} \multicolumn{9}{l}{\textit{Track 1}} \\
    Stand-In-14B~\cite{stand-in2025}   & 0.2396 & 0.6286 & - & 0.4878  & \textbf{0.4589} & 0.9112 & 0.5550 & 0.5042 \\
    RefAlign-14B~\cite{refalign2026}   & \textbf{0.7996} & 0.6342 & - & 0.4758 & 0.2946 & 0.9637 & 0.4837 & 0.6214 \\
    Phantom-14B~\cite{phantom2025}     & 0.7165 & 0.6242 & - & 0.5041 & 0.2102 & 0.8882 & 0.5712 & 0.6211 \\
    IPT2V-14B~\cite{wang2025identity}  & 0.5387 & 0.5572 & - & 0.4940 & 0.3864 & 0.9004 & 0.5387 & 0.5614 \\
    \textbf{USTC-CMI(Chronos)}           & 0.6840 & \textbf{0.6411} & - & \textbf{0.5984} & 0.4306 & \textbf{0.9709} & \textbf{0.6725} & \textbf{0.6769} \\
    
    \midrule %
    
    \rowcolor{gray!15} \multicolumn{9}{l}{\textit{Track 2}} \\
    Stand-In-14B~\cite{stand-in2025}   & 0.1844 & \textbf{0.7268} & \textbf{0.5663} & 0.4836 & \textbf{0.3027} & 0.9191 & 0.6350 & 0.5100 \\
    RefAlign-14B~\cite{refalign2026}   & 0.3877 & 0.6996 & 0.5392 & 0.4930 & 0.1620 & 0.9838 & 0.5812 & 0.5430 \\
    Phantom-14B~\cite{phantom2025}     & 0.4136 & 0.7170 & 0.5432 & 0.5268 & 0.2110 & 0.9613 & 0.6387 & 0.5725 \\
    \textbf{USTC-CMI(Chronos)}           & \textbf{0.4400} & 0.6907 & 0.5407 & \textbf{0.6009} & 0.1479 & \textbf{0.9950} & \textbf{0.70625} & \textbf{0.6102} \\
    \bottomrule
  \end{tabular}
  }
\end{table*}
Given the enhanced prompt $\hat{P}$ and VFA-derived visual anchors $\mathcal{A}$, we introduce \textbf{Metric-Guided Candidate Routing} (MGCR) to automatically identify the most reliable video from a heterogeneous generator pool based on multi-dimensional evaluation.
With a set of video generators $\mathcal{G}=\{G_k\}_{k=1}^{K}$, we produce a candidate set:
\begin{equation}
    \mathcal{V}=\{V_k \mid V_k=G_k(\hat{P},\mathcal{A}),\; G_k\in\mathcal{G}\}.
\end{equation}

Rather than relying on a single generator, this design leverages the complementary capabilities of different models in identity preservation, prompt adherence, and visual fidelity.
For candidate routing, we employ an automatic evaluation system that measures each video across complementary criteria, including text consistency, multi-subject identity preservation, motion amplitude, visual quality, and LLM-based holistic assessment~\cite{yuan2026opens2v}.
The evaluator maps each candidate video $V_k$ to a normalized score vector:
\begin{equation}
    \mathbf{s}(V_k)=
    \big[
    s_{\mathrm{text}},
    s_{\mathrm{id}},
    s_{\mathrm{subj}},
    s_{\mathrm{motion}},
    s_{\mathrm{qual}},
    s_{\mathrm{llm}}
    \big],
\end{equation}
where each dimension captures a complementary aspect of generation fidelity. 
We then compute the final routing score:
\begin{equation}
    R(V_k)=\sum_{j}\lambda_j s_j(V_k),
\end{equation}
and the highest-scoring output is selected as:
\begin{equation}
    V^{*}=\arg\max_{V_k\in\mathcal{V}} R(V_k).
\end{equation}

To improve robustness, we further introduce an \textbf{exception-aware review} mechanism. A selected candidate is not directly accepted if it exhibits an abnormally low score in any critical dimension, even when its aggregated routing score remains high.
Formally, we assign the selected result to the human review set whenever it violates any critical metric constraint:
\begin{equation}
    \exists j\in\mathcal{J}_{\mathrm{crit}},\quad s_j(V^{*})<\tau_j,
\end{equation}
where $\mathcal{J}_{\mathrm{crit}}$ denotes the set of critical metrics and $\tau_j$ is the corresponding failure threshold. 
This mechanism retains the efficiency of automatic routing while reducing the risk of severe failure cases being silently selected as final submissions.

\section{Experiments}
\begin{table*}[t]
  \centering
  \caption{Quantitative ablation studies on each component on Track 1 and Track 2. Chronos achieves the highest Total Score in both tracks, validating the individual and synergistic contributions of each module toward improving overall video quality. }
  \label{tab:ablation_results_merged}
  \resizebox{\linewidth}{!}{
  \begin{tabular}{ccccccccc}
    \toprule
    \multirow{2}{*}{\textbf{Method}} &
    \textbf{Identity Consistency} &
    \multicolumn{2}{c}{\textbf{Text Alignment}} &
    \multicolumn{4}{c}{\textbf{Video Quality}} &
    \multirow{2}{*}{\textbf{Total Score$\uparrow$}} \\
    \cmidrule(lr){2-2}  \cmidrule(lr){3-4} \cmidrule(lr){5-8} 
    & \textbf{\makecell{Subject Score$\uparrow$}}
    & \textbf{\makecell{GMEScore$\uparrow$}}
    & \textbf{\makecell{GMEScore-Seg$\uparrow$}}
    & \textbf{\makecell{Aesthetic Score$\uparrow$}}
    & \textbf{\makecell{Motion Amplitude$\uparrow$}}
    & \textbf{\makecell{Motion Smoothness$\uparrow$}}
    & \textbf{\makecell{Natural Score$\uparrow$}} & \\
    \midrule
    
    \rowcolor{gray!15} \multicolumn{9}{l}{\textit{Track 1}} \\
    baseline            & 0.7511 & 0.6188 & - & 0.5001 & 0.1827 & 0.9548 & 0.5775 & 0.6353 \\
    + SPA            & \textbf{0.7557} & 0.6176 & - & 0.5030 & 0.1763 & 0.9529 & 0.5875 & 0.6394 \\
    + SPA\ \&\  VFA (use Seedance)          & 0.6736 & 0.6357 & - & 0.5824 & \textbf{0.4996} & 0.9645 & 0.6475 & 0.6646 \\ %
    + SPA\ \&\  VFA (use Wan)          & 0.6816 & \textbf{0.6650} & - & 0.6000 & 0.3652 & 0.9508 & 0.5800 & 0.6487   \\
    \textbf{USTC-CMI(Chronos)} & 0.6981 & 0.6475 & - & \textbf{0.6063} & 0.4798 & \textbf{0.9762} & \textbf{0.6675} & \textbf{0.6833} \\
    
    \midrule %
    
    \rowcolor{gray!15} \multicolumn{9}{l}{\textit{Track 2}} \\
    baseline            & 0.4576 & \textbf{0.6872} & \textbf{0.5504} & 0.5278 & 0.1893 & 0.9632 & 0.6300  & 0.5803 \\
    + SPA\ \&\  VFA (use Wan I2V-14B)         & 0.5338 & 0.6741 & 0.5502 & 0.6131 & \textbf{0.1963} & 0.9825 & 0.6450 & 0.6177 \\
    + SPA\ \&\  VFA (use Wan FLF2V-14B)          & 0.5911 & 0.6771 & 0.5473 & \textbf{0.6439} & 0.1505 & \textbf{0.9956} & 0.6625 & 0.6417  \\
    \textbf{USTC-CMI(Chronos)} & \textbf{0.6085} & 0.6761 & 0.5496 & 0.6406 & 0.1743 & 0.9944 & \textbf{0.6850} & \textbf{0.6530} \\
    \bottomrule
  \end{tabular}
  }
\end{table*}
\begin{figure*}[t]
    \centering
    \includegraphics[width=\linewidth]{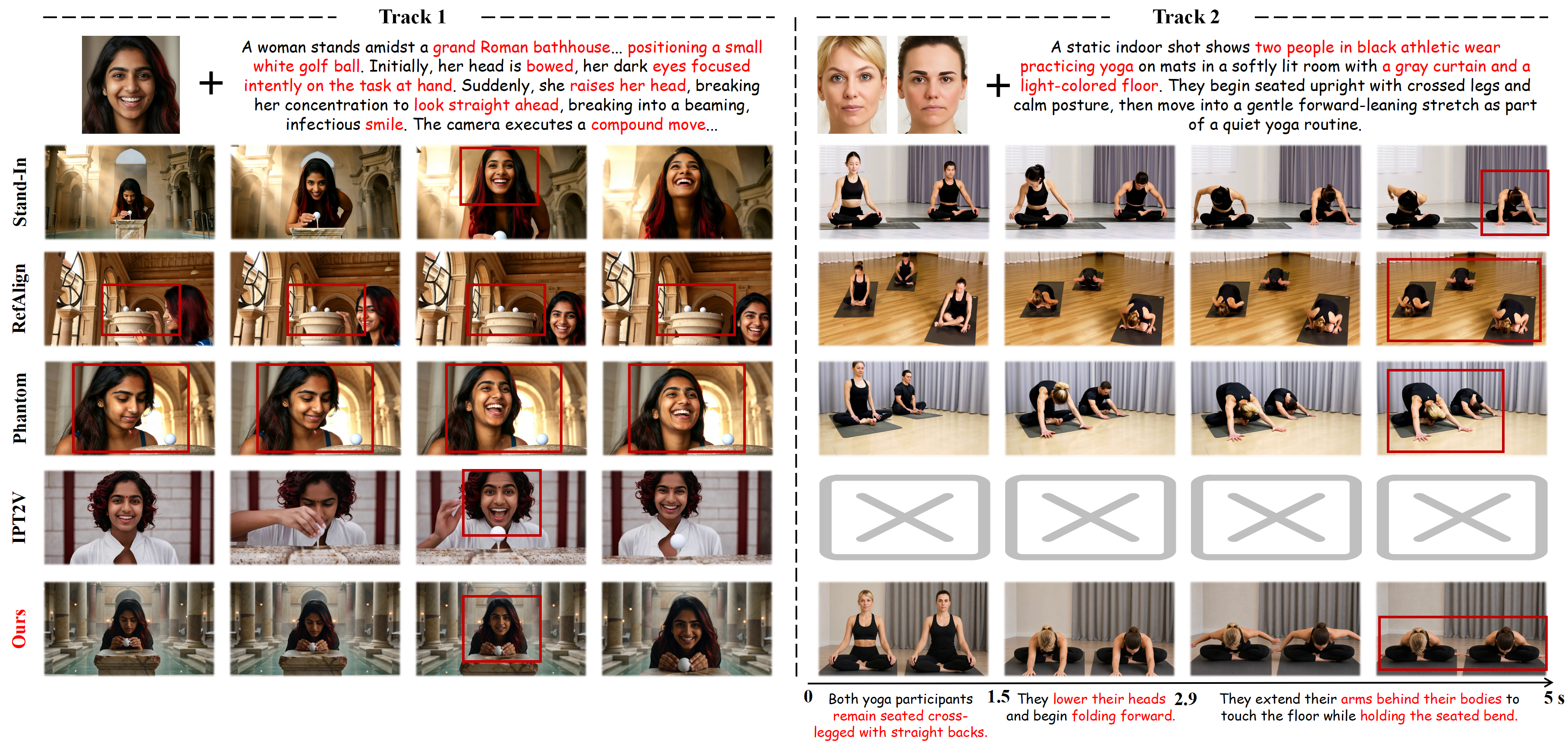}
    \caption{\textbf{Visualization results of different methods on Track 1 (left) and Track 2 (right)}. For Track 1, while Stand-In and IPT2V generate actions corresponding to the prompt, they suffer from low identity consistency; conversely, RefAlign and Phantom maintain better identity consistency but fail to perform the prompt-specified action (e.g., placing a golf ball). In contrast, \textit{Chronos} achieves high identity fidelity while accurately executing the corresponding motion. For Track 2, \textit{Chronos} demonstrates robust responsiveness to sequential actions and achieves superior temporal and multi-reference identity consistency.}
    \Description{A side-by-side qualitative comparison for two challenge tracks. On the left, Track 1 shows a reference portrait and a golf-ball prompt above frame sequences from Stand-In, RefAlign, Phantom, IPT2V, and Chronos; red boxes highlight identity or action differences, while the Chronos sequence preserves the woman's identity as she places the ball and raises her head to smile. On the right, Track 2 shows two reference portraits and a sequential yoga prompt above corresponding method outputs; the Chronos sequence preserves both participants while they progress from seated poses through forward bends over five seconds. Gray crossed boxes indicate unavailable IPT2V results, red text marks key requested actions, and red boxes identify regions emphasized for comparison.}
    \label{fig:main_results}
\end{figure*}
\subsection{Setup}
\textbf{Challenge Setup.}
Identity-Preserving Video Generation (IPVG) task aims to maintain the consistency of the given reference identity along the text-to-video generation process. This year's IPVG grand challenge includes two tracks: Facial Identity-Preserving Video Generation (Track1) and Sequential Action Identity-Preserving Video Generation (Track2). The challenge website is \url{https://hidream-ai.github.io/ipvg-challenge-2026.github.io}

\noindent\textbf{Models.} We instantiate the generator pool with Wan2.2 I2V-14B~\cite{wan2025wan} for both Track 1 and Track 2, complemented by Seedance~\cite{seedance2026seedance} for Track 1 and Wan FLF2V~\cite{wan2025wan} for Track 2. 
Both SPA and VFA are implemented with GPT-based models~\cite{openai_gpt5_4, openai_chatgpt_images_2}.

\noindent\textbf{Baseline.}
To validate the effectiveness of our method, we compare \textit{Chronos} on the IPVG task with state-of-the-art methods, including Stand-In~\cite{stand-in2025}, RefAlign~\cite{refalign2026}, and Phantom~\cite{phantom2025}. In addition, we also compare with IPT2V~\cite{wang2025identity} on Track 1.

\noindent\textbf{Metrics.}
We use an automatic evaluation system to score generated videos across three dimensions:
Identity is evaluated using FaceSim-Cur~\cite{deng2019arcface} for humans and YOLO-World-based crop similarity~\cite{cheng2024yolo} for other subjects. Text alignment uses GMEScore~\cite{yuan2026opens2v}. Video quality uses Aesthetic Score(AS)~\cite{schuhmann2024christophschuhmann}, Motion Amplitude(MA)~\cite{bradski2000opencv}, Motion Smoothness(MS)~\cite{ling2025vmbench}, and Natural Score(NS)~\cite{yuan2026opens2v}.

\subsection{Quantitative Results}
As illustrated in Tab.~\ref{tab:challenge_results}, the proposed \textbf{\textit{Chronos} achieves first place in both Track 1 and Track 2 of the ACM MM 2026 Identity-Preserving Video Generation Challenge}.

\textit{Chronos} achieves the highest Total Score in both Track 1 and Track 2, demonstrating its superior overall performance, as presented in Tab.~\ref{tab:main_results}.
In Track 1, \textit{Chronos} outperforms the runner-up RefAlign by approximately 9.0\% in Total Score. While RefAlign yields higher facial similarity, it compromises visual quality, motion smoothness, and naturalness. Conversely, \textit{Chronos} strikes a superior balance, maintaining strong facial fidelity while ranking first in Text Alignment, AS, MS, and NS. 
In Track 2, \textit{Chronos} improves the Total Score by 6.6\% over the runner-up \textit{Phantom}, securing the highest Subject Score for multi-reference consistency across diverse entities, while also ranking first in key visual quality metrics.

\subsection{Qualitative Results}

Qualitative experiments in Fig.~\ref{fig:main_results} show that while Stand-In and IPT2V generate actions corresponding to the prompt, they suffer from low identity consistency. Conversely, Phantom maintains better identity consistency but fails to perform the prompt-specified action, whereas RefAlign suffers from poor text alignment and fails to capture the quantity relationships of the subjects. In contrast, \textit{Chronos} achieves high identity fidelity while accurately executing the corresponding motion. Furthermore, owing to the introduction of the proposed VFA module, \textit{Chronos} yields more aesthetically pleasing and highly controllable visual layouts.

\subsection{Ablation Study} 
Due to the high evaluation cost, we conduct ablation studies on a representative subset of the first 100 cases, where the observed trends remain consistent with those of the full evaluation set.
Using VACE~\cite{vace2025} as the baseline, we progressively evaluate the effectiveness of each proposed strategy.
Tab.~\ref{tab:ablation_results_merged} reports the ablation results of our core modules on both Track 1 and Track 2.
First, SPA consistently strengthens identity preservation, leading to improved Subject Score on Track 1.
Second, VFA further improves visual quality across different generator backbones, including Seedance and Wan variants, with notable gains in AS and NS.
\textit{Chronos} achieves the strongest performance on both tracks, validating the effectiveness and complementary benefits of the proposed components.

\section{Conclusion}
In this paper, we presented \textit{Chronos}, a novel training-free framework for identity-preserving video generation that unifies single-segment facial identity preservation and multi-segment sequential action generation within a common temporal formulation. \textit{Chronos} first extracts compact identity cues from reference images to adaptively adjust visibility and viewpoint cues in the textual prompt, ensuring identity fidelity without compromising the original scene semantics or motion intent. It then transforms these enhanced conditions into visually and temporally grounded anchors in a cascaded manner, reducing the ambiguity of text-only conditioning and providing robust guidance for identity preservation under challenging pose variations and complex action transitions. Finally, an automatic multi-dimensional evaluation system scores and selects optimal outputs from heterogeneous video generators, leveraging their complementary strengths while incorporating an exception-aware mechanism to filter critical generation failures. Extensive evaluations demonstrate the superiority of \textit{Chronos}, which achieved first place in both Track 1 and Track 2 of the ACM MM 2026 Identity-Preserving Video Generation Challenge.

\begin{acks}
This research is supported by the Artificial Intelligence National Science
and Technology Major Project 2023ZD0121200, and the National Natural Science
Foundation of China under Grant 623B2094.
\end{acks}

\FloatBarrier
\bibliographystyle{ACM-Reference-Format}
\bibliography{references}

\end{document}